\documentclass{article}
\usepackage[utf8]{inputenc}
\usepackage{color}
\usepackage{amsmath}
\usepackage{setspace}
\usepackage{graphicx}
\usepackage{pgfplots}
\usepackage{authblk}

\title{Return of the RNN: Residual Recurrent Networks for Invertible Sentence Embeddings}
\author{Jeremy Wilkerson\thanks{jeremy@talkingape.org}}
\affil{talkingape.org}
\date{April 5, 2023}

\begin{document}
	\maketitle
	\doublespacing
	
	\section*{Abstract}
	This study presents a novel model for invertible sentence embeddings using a residual recurrent network trained on an unsupervised encoding task. Rather than the probabilistic outputs common to neural machine translation models, our approach employs a regression-based output layer to reconstruct the input sequence's word vectors. The model achieves high accuracy and fast training with the ADAM optimizer, a significant finding given that RNNs typically require memory units, such as LSTMs, or second-order optimization methods. We incorporate residual connections and introduce a ``match drop'' technique, where gradients are calculated only for incorrect words. Our approach demonstrates potential for various natural language processing applications, particularly in neural network-based systems that require high-quality sentence embeddings.
	
	\section{Introduction}
	The concept of representing words as continuous vectors has its roots in the distributional hypothesis (Harris, 1954), which posits that words with similar meanings tend to occur in similar contexts. In recent years, several pre-trained word embedding models, such as Word2Vec (Mikolov et al., 2013a, b) and GloVe (Pennington et al., 2014), have been developed to capture these contextual similarities. These high-dimensional vector representations capture semantic and syntactic relationships, proving useful for various natural language processing (NLP) tasks.
	
	Moving beyond word embeddings, researchers have focused on creating fixed-length representations for entire sentences. Methods such as Skip-Thought Vectors (Kiros et al., 2015) and Quick-Thought Vectors (Logeswaran and Lee, 2018) have been proposed to learn general-purpose sentence embeddings using an encoder-decoder framework, building upon advances in sequence-to-sequence (seq2seq) models (Sutskever et al., 2014). Seq2seq models and their variants, including those with attention mechanisms (Bahdanau et al., 2015), have been widely adopted for tasks like machine translation and text summarization.
	
	Recurrent Neural Networks (RNNs) have played a pivotal role in the development of sequence modeling and NLP tasks, particularly in the context of seq2seq models. RNNs excel at capturing temporal dependencies in sequential data, but traditional RNNs suffer from the vanishing gradient problem. Gated variants, such as Long Short-Term Memory (LSTM) networks (Hochreiter and Schmidhuber, 1997) and Gated Recurrent Units (GRUs) (Cho et al., 2014), have been developed to address this issue, improving RNN performance in various sequence modeling tasks.
	
	Residual networks (ResNets) have also been a significant breakthrough in deep learning, addressing the vanishing gradient problem for very deep networks (He et al., 2016a). ResNet-v2, an improved version of the original ResNet, introduced identity mappings that further enhance the flow of information and gradients during training (He et al., 2016b). ResNets have been adopted in various deep learning applications, including NLP.
	
	As NLP models become more complex and generate larger embeddings, compression techniques have become increasingly important. Methods such as principal component analysis (PCA) and autoencoders have been employed to compress high-dimensional embeddings while preserving essential information (Hinton and Salakhutdinov, 2006).
	
	In this study, we introduce a novel model for invertible sentence embeddings that employs a residual recurrent network and our newly proposed match drop technique. Our approach builds upon the advancements in word embeddings, seq2seq models, RNNs, and ResNets, demonstrating potential for various NLP applications, including neural network-based systems that require high-quality sentence embeddings.

	\section{Methodology}
	
	\subsection{Model Architecture}
\begin{figure}[htbp]
		\centering
		\includegraphics[width=0.9\textwidth]{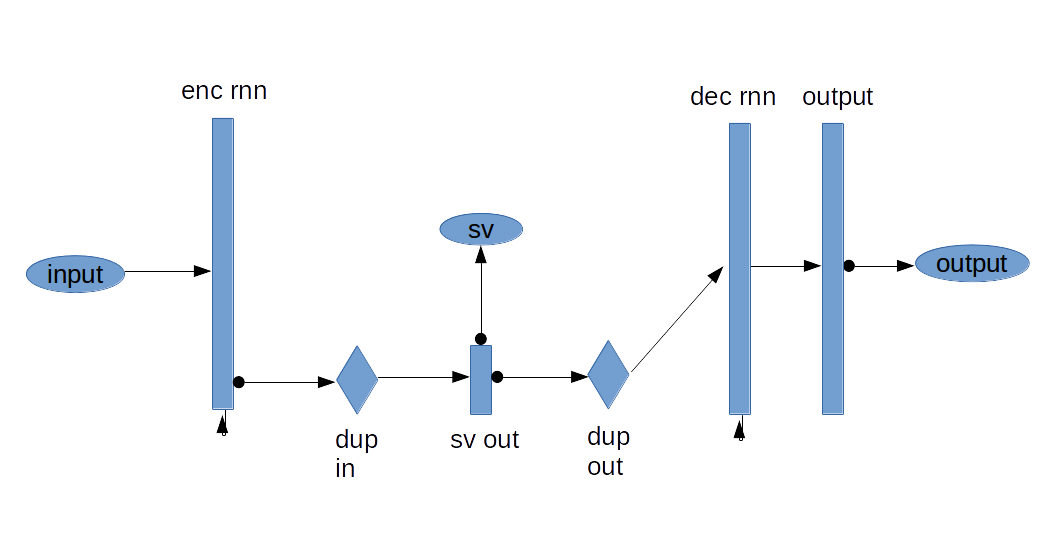}
		\caption{The encoder/decoder network architecture. The input and output are sequences of word vectors, of size 300 each. \emph{enc rnn} and \emph{dec rnn} are residual recurrent layers. \emph{dup in} represents the operation of taking the RNN output at the last time step. \emph{dup out} represents the operation of repeating the vector at each time step. \emph{sv out} and \emph{output} are fully-connected layers. All arrows without circles at their bases represent fully-connected inputs.}
		\label{fig:model}
	\end{figure}

	Our model represents sentences as sequences of word vectors, which are derived from pre-trained 300-dimensional Word2Vec embeddings. The model contains an encoder that encodes the sequence of word vectors into a sentence vector and a decoder that decodes the sentence vector back into a sequence of word vectors. In contrast to conventional neural machine translation (NMT) models with probabilistic outputs, the output layer in our model utilizes regression to reconstruct the input sequence of word vectors. Refer to Figure \ref{fig:model} for a visual representation of the network architecture. A single fully-connected layer, equal in size to the encoder and decoder, is situated between the encoder and decoder, which was discovered to enhance training and model accuracy. The hyperbolic tangent (tanh) function serves as the nonlinear activation function throughout the model, but no activation function is used on the output layer. Dropout (Srivastava et al., 2014) is applied to the outputs of both the encoder and decoder layers, utilizing a dropout rate of 0.5 in each case. 
	
	Besides the encoder/decoder network, an independent compressor/decompressor network is also incorporated in the model.

	The model is implemented and executed using custom C++ CUDA code to take advantage of GPU acceleration and optimize performance during training and inference.

	\subsection{Loss Function and Match Drop}
	In our study, we employed a loss function that incorporates squared error along with a ``match drop'' technique. This technique omits words that already match the expected output from the gradient calculation, allowing the model to focus its computational resources on the remaining mismatched words.
	
	The match drop technique facilitates more efficient training by prioritizing words that are harder to match. Some words are matched more easily, such as with a cosine similarity of around 0.5, while others demand higher similarity for accurate matching. Some examples of word types that require a higher similarity to match include digits (excluding 0 and 1), names of months or days of the week, and proper nouns. Without match drop, the network would achieve an extremely high similarity, nearly 1, for common words, at the expense of matching less frequent words or words that are harder to match. With match drop, the network will not try to improve the similarity on words that are already matching.
	
	The match drop technique can be conceptualized as follows: for each word there is a volume in the word embedding space that consists of all points closer to that word than to any other, with the word vector at the ``center'' of that volume. The loss function can then be defined as a piecewise function:	
	
	\begin{equation}
	Loss =
		\begin{cases}
			\text{distance from word center} & \text{if output vector is outside the word volume} \\
			0 & \text{if output vector is inside the word volume}
		\end{cases}
	\end{equation}

	This equation indicates that if the output is not a match, the loss corresponds to the distance between the output vector and the target word vector. On the other hand, if there is a match, the loss is 0.

	By incorporating the match drop technique into the loss function, we ensure that the model efficiently allocates its computational power, improving the overall performance and preserving the semantic information of the input sentences.
	
	\subsection{Matching}
	Matches were identified by locating the word in the Word2Vec dictionary that exhibits the highest cosine similarity with the output vector. To efficiently compute these matches, we performed matrix multiplication between the output vectors and the Word2Vec dictionary. This process enabled us to simultaneously compare the output vectors with all words in the dictionary, allowing for a fast and effective calculation of the closest matching words.
	
	At first glance, it might appear counter-intuitive that the loss function employs squared error while matching is determined using cosine similarity. However, the use of cosine similarity for matching is essential to ensure efficient computation. Experiments were conducted to train the network using cosine similarity as the loss function, in order to align it with the match calculation process. Surprisingly, these experiments resulted in lower accuracy, indicating that employing squared error in the loss function was indeed more effective.
	
	\subsection{Encoder and Decoder Layers}
\begin{figure}[htbp]
		\centering
		\includegraphics[width=0.9\textwidth]{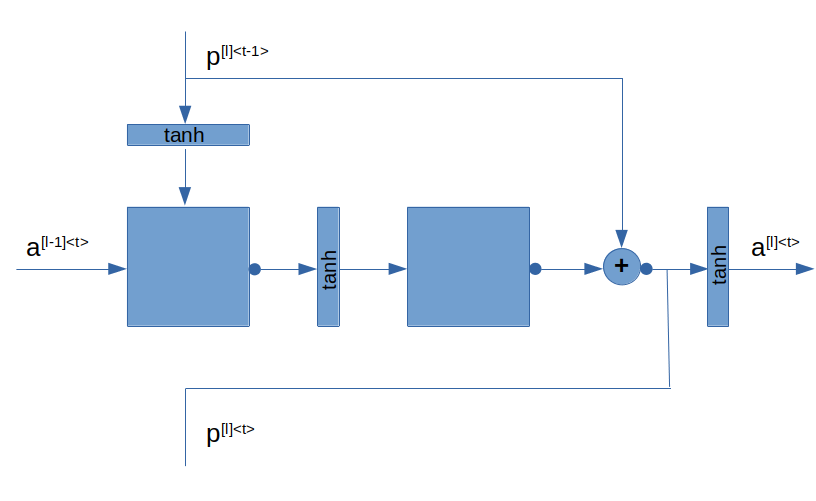}
		\caption{The residual RNN architecture. The arrows showing as inputs to the two square boxes are fully-connected inputs. The two inputs to the first box are concatenated. The circle with + represents addition. $a^{[l-1]<t>}$ represents the activation input from the previous layer and $a^{[l]<t>}$ represents the activation output to the next layer. $p^{[l]<t-1>}$ represents the pre-activation recurrent input from the previous time step, and $p^{[l]<t>}$ represents the pre-activation recurrent output to the next time step.}
		\label{fig:rrnn}
	\end{figure}

	The recurrent encoder and decoder layers in our model employ simple neurons rather than special memory cells such as LSTM or GRU units. These layers feature a residual architecture, similar to ResNet-v2 (He et al., 2016b), with an identity mapping and a residual calculated from two fully-connected layers at each time step, as illustrated in Figure \ref{fig:rrnn}. Unlike ResNet-v2, our network does not incorporate batch normalization layers. In the encoder layer, the output from the final time step serves as the overall layer output. Meanwhile, the decoder layer takes the same sentence vector input at each time step and generates an output word for every step. A bias vector is used as the recurrent input at the first time step for both layers. Notably, the encoder processes sentences in reverse order, while the decoder generates its output as a sentence in the standard, forward order.

	\subsection{Training}
	We trained the model, which consists of 606,070,300 parameters, using the ADAM algorithm. We stopped training when the number of matched words on the test dataset was no longer increasing significantly, which required 2,220,000 iterations. With a minibatch size of 32, there were 71,040,000 sentences processed, with no repetitions.
	
	The model was optimized using an initial learning rate of $4.22 \times 10^{-5}$. To adjust the learning rate during training, we applied a decay factor of 0.9999987 per iteration. We employed L2 regularization with a coefficient of $1.84 \times 10^{-7}$. We used ADAM p1 and p2 coefficients of 0.85 and 0.99, respectively. To tune our model's hyperparameters, we conducted a random hyperparameter search.

	The training process, which spanned 21 days, was executed on a hardware setup consisting of an NVIDIA GeForce RTX 3090 GPU with 24 GB of video memory and an Intel(R) Core(TM) i7-10700 CPU @ 2.90GHz. Training the model utilized less than 11 GB of video memory due to the use of mixed-precision floating-point arithmetic. Notably, the majority of the training time was dedicated to the calculation of matches for the match drop technique.
	
	\subsection{Compression}
	The independent compressor/decompressor network employed in our study is a straightforward yet effective design, comprising two fully-connected layers. The first layer, the compression layer, has a size of 3000 and is responsible for reducing the initial 10,000-dimensional sentence vector into a more compact representation. The second layer, the decompression layer, has a size of 10,000 and reconstructs the compressed 3000-dimensional vector back to its original size. This network is trained with ADAM as an autoencoder, utilizing the compression and decompression layers to efficiently encode and decode the sentence vector while preserving the essential information contained within it. It required less than a day to train on the aforementioned hardware setup. There is a potential for higher compression ratios using more complex networks or algorithms.
	
	\subsection{Data Preprocessing}
	For this study, we utilized the WMT datasets from 2007 to 2017 (Bojar et al., 2017), which contain individual sentences without any specific order. We began by consolidating the English datasets from each year and eliminating duplicate entries.
	
	During the preprocessing stage, we implemented a series of steps to prepare the data for training. Initially, we standardized non-ASCII characters by converting them to ASCII equivalents whenever possible. Any remaining non-ASCII characters were subsequently removed from the dataset.
	
	We then tokenized the sentences using the Stanford CoreNLP natural language processing toolkit (Manning et al., 2014).
	
	In order to maintain consistency within the dataset and ensure accurate evaluation of the model's performance, non-English sentences were identified and removed using heuristics. This step was taken to reduce noise in the dataset and focus on the model's ability to process English text effectively.
	
	Subsequently, we standardized all punctuation to a limited set.
	
	Next, we identified instances of multi-word phrases within the text that were present in the Word2Vec dictionary. In these cases, we combined the individual words into a single multi-word phrase. For instance, ``Abraham Lincoln'' was replaced by ``Abraham\_Lincoln''. This approach shortened sentences, enabling the representation of more information within a given sentence length. Additionally, it allowed for a more effective utilization of the semantic information encoded in the Word2Vec vectors.
	
	Finally, we converted numeric digits in the dataset to their corresponding word representations, such as replacing the digit `2' with the word `two', as this was found to enhance training and model accuracy.
	
	We divided the resulting set of unique, processed sentences into separate train, tune, and test datasets, allocating the vast majority to the train dataset.
	
	\subsection{Special Vectors}
	Randomly generated ``special vectors'' represented punctuation, common words missing from the word embeddings, end-of-sentence markers, and unknown-word markers. These vectors were designed to have similar statistics as the input word embeddings.

	\section{Results}
	We trained and tested our model using sentences with 60 words or fewer from the combined WMT dataset, with the mean sentence length being about 23 words. With a sentence vector size of 10,000, the model achieved an accuracy of 97\% of words matching, and 68\% of sentences reproduced exactly, on the test dataset. The compressor network reduced the 10,000-dimensional vector to 3,000 dimensions, maintaining nearly the same accuracy after decompression and decoding. 
	
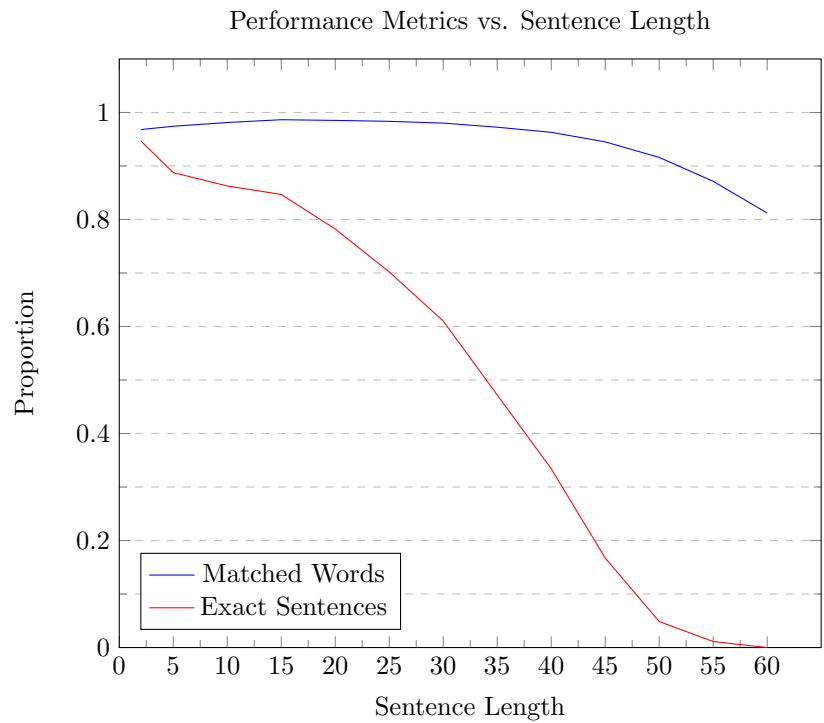
\begin{figure}
	\centering
	\begin{tikzpicture}
		\begin{axis}[
			width=0.9\textwidth,
			title={Performance Metrics vs. Sentence Length},
			xlabel={Sentence Length},
			ylabel={Proportion},
		    xmin=0, xmax=65,
			ymin=0, ymax=1.1,
			xtick={0,5,10,15,20,25,30,35,40,45,50,55,60},
			ytick={0,0.2,0.4,0.6,0.8,1.0},
			minor tick num=1,
			legend pos=south west,
			ymajorgrids=true,
			yminorgrids=true,
			grid style=dashed,
			]
	
			\addplot[
			color=blue,
			mark=none,
			]
			coordinates {
				(2,0.968085)(5,0.974241)(10,0.981263)(15,0.986469)(20,0.985184)(25,0.983443)(30,0.980166)(35,0.972496)(40,0.962926)(45,0.945075)(50,0.916054)(55,0.871399)(60,0.812079)
			};
			\addlegendentry{Matched Words}
			
			\addplot[
			color=red,
			mark=none,
			]
			coordinates {
				(2,0.946809)(5,0.887613)(10,0.862725)(15,0.846772)(20,0.782246)(25,0.702063)(30,0.610265)(35,0.472129)(40,0.334229)(45,0.167394)(50,0.0486486)(55,0.011236)(60,0)
			};
			\addlegendentry{Exact Sentences}
			
		\end{axis}
	\end{tikzpicture}
	\caption{Plot showing the relationship between sentence length and performance metrics (matched words and exact sentences).}
	\label{fig:metrics}
	\end{figure}

	Figure \ref{fig:metrics} shows plots of matched words and exact sentences verses sentence length.
	
	Table \ref{table:input_output_sentences} shows, for 10 sentences chosen randomly from the test data, the original sentence and the output sentence resulting from encoding and decoding. The mismatched words, of which there are only 2, are shown in red.

	\begin{table}[h!]
		\centering
		\begin{tabular}{|p{0.45\linewidth}|p{0.45\linewidth}|}
			\hline
			\textbf{Input Sentence} & \textbf{Output Sentence} \\ \hline
			" They know what they are buying EOS & " They know what they are buying EOS \\ \hline
			Health.com : The 1 0 best workouts for your sex life EOS & Health.com : The 1 0 best workouts for your sex life EOS \\ \hline
			A line of emergency vehicles arrives near the scene of the explosion EOS & A line of emergency vehicles arrives near the scene of the explosion EOS \\ \hline
			Thanks to skills fostered by her father , she decides to open the only female - owned detective agency in Botswana EOS & Thanks to skills fostered by her father , she decides to open the only female - owned detective agency in Botswana EOS \\ \hline
			Despite the fact that she was raised in a palace with 3 0 servants India s culture then she said was not one of constant acquisition : " We had a lot of money but there wasn't anything to buy " EOS & Despite the fact that she was raised in a palace with 3 0 servants India s culture then she said was not one of constant acquisition : " We had a lot of money but there wasn't anything to buy " EOS \\ \hline
			Outside the room , a devastated Beshara said : " You'd hate to see them in charge of the death penalty " EOS & Outside the room , a devastated \textcolor{red}{Bragunier} said : " You'd hate to see them in charge of the death penalty " EOS \\ \hline
			His truck s refrigerator is stocked with chicken , tuna and vegetables EOS & His truck s refrigerator is stocked with chicken , \textcolor{red}{pimientos} and vegetables EOS \\ \hline
			Thankfully we got the goal and then finished the game well EOS & Thankfully we got the goal and then finished the game well EOS \\ \hline
			Costa received a text from Chelsea manager Antonio\_Conte saying he was no longer in his plans and has set his mind on a return to former club Atletico who are under a transfer embargo EOS & Costa received a text from Chelsea manager Antonio\_Conte saying he was no longer in his plans and has set his mind on a return to former club Atletico who are under a transfer embargo EOS \\ \hline
			We're playing by the rules that exist EOS & We're playing by the rules that exist EOS \\ \hline
		\end{tabular}
		\caption{Example Input and Output Sentences with Mismatched Words Highlighted. EOS represents the end-of-sentence marker.}
		\label{table:input_output_sentences}
	\end{table}

	\section{Discussion}
	In this study, we proposed a model for invertible sentence embeddings using a residual recurrent network trained for regression on an unsupervised encoding task. The model demonstrates several key advantages and novel contributions to the field of natural language processing.
	
	Firstly, the ability to train the residual recurrent network with the ADAM optimizer is significant, given the known challenges associated with training vanilla RNNs. As we discussed earlier, RNNs typically require memory units, such as LSTMs, or second-order optimization methods to overcome issues such as the vanishing gradient problem. However, our model leverages residual connections and a match drop technique to achieve high performance and fast training without relying on special memory units or second-order optimization methods. This finding highlights the potential benefits of residual architectures in addressing the limitations of vanilla RNNs and suggests the possibility of applying such architectures to other RNN-based tasks and models.
	
	The match drop technique, introduced in this study, constitutes a novel contribution to the field. By selectively focusing computational resources on words that are more challenging to match, the match drop technique improves overall training efficiency and effectiveness. It allows the network to concentrate on refining its predictions for less frequent and more semantically complex words, enhancing the model's ability to capture and preserve the semantic information contained in sentence embeddings. This innovative approach has the potential to significantly impact the field of neural machine translation and beyond, opening new avenues for further research and optimization in neural network training methodologies.
	
	Furthermore, our study demonstrates that the proposed model is capable of achieving high fidelity in the encoding and decoding of sentences, with a 97\% matching accuracy of exact words. In cases where the word did not match, it was often a synonym or near-synonym, indicating that a system using the output of our model may still be able to maintain the context and meaning of the original sentence. This suggests that the compressed sentence vector could serve as input to another neural network, with the output of that network being decompressed and decoded by our network to construct an output sentence with high accuracy.
	
	The ability to decode a sentence vector into its exact original form serves as a sufficient condition to establish that the vector encompasses all the semantic information present in the sentence. On the other hand, when a sentence cannot be reconstructed precisely, such as the 3\% mismatched words observed with our model, it signifies a loss of semantic information and diminishes the model's usefulness. However, a system that utilizes the output of our model might be able to leverage the context of a sentence to compensate for any lost semantic information, thus maintaining the overall meaning and coherence of the text. Our experiments indicate that increasing the size of the encoder, decoder, and sentence vector -- all of which share the same dimensions -- boosts reconstruction accuracy. Consequently, it should be possible to achieve an accuracy above 97\% by increasing the original vector size beyond 10,000. This adjustment could also require expanding the compressed vector size above 3,000 to preserve the reconstruction accuracy.
	
	The model had difficulty distinguishing between days of the week, months, and digits (excluding 0 and 1) within their respective categories due to the close proximity of their Word2Vec representations. Adjusting or fine-tuning these representations to create greater separation could potentially enhance the model's match rate, not only for these specific words but also for other words with the computational power that would be freed up as a result.

	\section{Conclusion}
	In this study, we have introduced a novel model for invertible sentence embeddings that employs a residual recurrent network and an innovative match drop technique we developed. Our model overcomes the limitations associated with training vanilla RNNs, achieving high performance and fast training using the ADAM optimizer without relying on special memory units or second-order optimization methods. The success of the model demonstrates the potential benefits of residual architectures in addressing RNN-based tasks and models.
	
	The proposed model achieves high fidelity in the encoding and decoding process, with a 97\% matching accuracy of exact words. As a result, the compressed sentence vector could serve as input to another neural network, with the output of that network being decoded by our network to construct an output sentence with high accuracy.
	
	Our model's success in capturing semantic and syntactic aspects of input sentences within vector representations opens up potential applications in various natural language processing tasks that require accurate and compact sentence representations.
	
	In conclusion, the proposed residual recurrent network for invertible sentence embeddings demonstrates significant advancements in the field of natural language processing. The model's ability to effectively capture and reconstruct sentence information within vector representations, combined with its efficient training and high performance, underlines the potential for further exploration and application of similar approaches in other natural language processing tasks.
	
	\section*{Acknowledgment}
	We would like to acknowledge the contribution of ChatGPT, a language model developed by OpenAI, for its assistance in generating text and providing insights during the writing process.

\end{document}